\documentclass[runningheads]{llncs}

\usepackage[numbers,sort&compress]{natbib}
\usepackage{graphicx}
\usepackage{amssymb}
\usepackage{amsmath}

\usepackage{algorithm}
\usepackage{algpseudocode}

\usepackage{color}

\begin{document}
\title{Message Passing Neural Networks for Hypergraphs\thanks{Supported by The Canada Research Chairs program.}}
%
%\titlerunning{Abbreviated paper title}
% If the paper title is too long for the running head, you can set
% an abbreviated paper title here
%
\author{Sajjad Heydari\and
Lorenzo Livi\orcidID{0000-0001-6384-4743}}
\authorrunning{S. Heydari and L. Livi}
% First names are abbreviated in the running head.
% If there are more than two authors, 'et al.' is used.
%
\institute{Computer Science, Univeristy of Manitoba, Winnipeg, Canada
\email{heydaris@myumanitoba.ca}}
\maketitle              % typeset the header of the contribution
\begin{abstract}
Hypergraph representations are both more efficient and better suited to describe data characterized by relations between two or more objects.
In this work, we present a new graph neural network based on message passing capable of processing hypergraph-structured data. We show that the proposed model defines a design space for neural network models for hypergraphs, thus generalizing existing models for hypergraphs.
We report experiments on a benchmark dataset for node classification, highlighting the effectiveness of the proposed model with respect to other state-of-the-art methods for graphs and hypergraphs.
We also discuss the benefits of using hypergraph representations and, at the same time, highlight the limitation of using equivalent graph representations when the underlying problem has relations among more than two objects.
\keywords{Graph Neural Network\and Hypergraph \and Message passing.}
\end{abstract}

\section{Introduction}

Graphs perfectly capture structured data, not only representing data points, but also relations amongst them. Recently, Graph Neural Networks have been employed as a tool to directly process graph data with huge success in various tasks, such as node classification in citation network labeling~\cite{kipf2016semi, velivckovic2017graph, atwood2016diffusion}, link predictions in recommender networks\cite{zhang2018link}, and state prediction in traffic prediction or weather forecasting via sensory networks\cite{zhang2018gaan, yan2018spatial, yu2017spatio, cui2019traffic}.

Graph Neural Networks have provided a significant reduction in numbers of trainable parameters in a machine learning model with input that could be represented as graphs, much like Convolutional Neural Networks have done for tensors. This reduction in parameters allows complex problems to be addressed with much smaller datasets. However there still remains a set of problems that could benefit from a change in representation, namely the hypergraph representation.

Hypergraphs constitute a natural generalization of graphs, where relations are not restricted to represent the interaction of two objects: hypergraphs can encode higher-order relations, i.e. relations between two or more objects. Formally, a hypergraph $H=(V, \mathcal{E})$ is composed by a set of vertices, $V$, and a set of hyperedges, $\mathcal{E}\subset\mathcal{P}(V)$, where $\mathcal{P}(V)$ is the power set of $V$.
Accordingly we may refer to the vertices composing a hyperedge as $v\in e, e\in \mathcal{E}$.
In a citation network, for example, there are pair-wise relations between cited work and citing work. However, there are also bigger relations among multiple works that have been cited together in a single publication, which require hyperedges to be properly modeled.

Higher-order relations could still be encoded in a graph by performing suitable transformations (e.g. using line graphs). However, we find that none of the currently used encodings are as effective as directly processing hypergraphs. We note that well-established graph neural networks \cite{gilmer2017neural} fail to apply to hypergraphs due to their more complex structure. Therefore, newer methods for processing hypergraphs with neural networks have been introduced in the literature \cite{feng2019hypergraph}.

Here we introduce the Hypergraph Message Passing Neural Network (HMPNN), a Message Passing Neural Network model \cite{gilmer2017neural} that can process hypergraph-structured data of variable size. The main contributions of this paper can be summarized as follows:
\begin{itemize}
    \item We develop a new message passing neural network model for hypergraph-structured data;
    \item We show that hypergraph convolution methods from the literature can be seen as a special case of HMPNN;
    \item We show that the proposed model significantly outperforms other existing methods from the literature on a common node classification benchmark.
\end{itemize}

The reminder of this paper is structured as follows.
Section \ref{sec:related_work} contextualizes the paper.
In Section \ref{sec:graph_expansions}, we show that common approaches for transforming a hypergraph into a graph inevitably lead to loss of structural information, hence preventing the correct functioning of any machine learning method operating on them.
Section \ref{sec:hmpnn} presents the main contribution of this paper, namely a novel hypergraph neural network model.
In Section \ref{sec:experiments} we show experimental results on a common hypergraph benchmark used in the literature. Finally, Section \ref{sec:conclusions} concludes the paper.

\section{Related work}
\label{sec:related_work}

\citet{gilmer2017neural} reformulated existing graph neural networks in a framework called Message Passing Neural Networks. There are two phases in MPNN, the message passing and readout, which corresponds to convolutions and pooling operations, respectively. The convolutional layer consists of updating the hidden state of node $V$ at layer $t$ ($h^t_v$), based on the previous state of $v$ and their neighbors. The model reads:
\begin{equation}
\begin{aligned}
m_v^{t+1} &= \sum_{w\in N(v)} M_t(h^t_v, h^t_w, e_{vw}) \\
h^{t+1}_v &= U_t(h^t_v, m^{t+1}_v)
\end{aligned}
\end{equation}

The function $M$ prepares messages from each node to each of its neighbours, which are then aggregated together to form the incoming message for each node. $M$ can utilize the sender's features, the receiver features, or the weight of the edge between them. The incoming message is then processed with function $U$ along with the internal feature, to form the next feature of each node.

\subsection{Graph design space}
\label{sec:graph_design_space}

\citet{you2020design}
introduced a design space for graph neural networks. Their contributions involves a similarity metric and an evaluation method for GNNs. Their design space consists of three parts: intra-layer, inter-layer, and training configuration.

The intra-layer is responsible for creation of various GNN convolutions, while also considering dropout and batch normalization. It is described as follows:
\begin{equation*}
h_v^{(k+1)} = AGG({ACT(DROPOUT(BN(W^{(k)}h_u^{(k)} + b^{(k)}))), u \in \mathcal{N}(v)})
\end{equation*}

The inter-layer is responsible for the interconnection of the various layers. These includes choices for \textit{layer connectivity}, \textit{pre-process layers}, \textit{mp-layers} and \textit{post-process layers}; with layer connectivity having options such as \textit{stack}, \textit{skip-sum} and \textit{skip-cat}.

Finally, training configuration includes choices for batch size, learning rate, optimizer and number of epochs.

\subsection{Hypergraph neural networks and hypergraph representation}

The literature on graph neural networks for hypergraph-structured data is sparse.
\citet{jiang2019dynamic} introduce dynamic hypergraph neural networks, a neural network that both updates the structure of the underlying hypergraph as well as performing a two phase convolution, a vertex convolution and a hyperedge convolution. 
\citet{feng2019hypergraph} introduce hypergraph neural networks, a spectral convolution operation on hypergraphs, that considers weights of the hyperedges as well as the incident matrix.
\citet{yi2020hypergraph} introduce a method to perform hypergraph convolutions with recurrent neural networks.
\citet{chien2021you} introduce an extension to message passing operation for hypergraphs based on composition of two multiset functions.

\section{Graph expansions and loss of structural information}
\label{sec:graph_expansions}

Here, we consider the process of mapping hypergraph representations into graph representations.
There are various ways of encoding hyperedges involving three or more vertices in graphs.
We consider some popular approaches and show that they all have drawbacks that lead to loss of relevant structural information, thus highlighting the importance of processing hypergraph-structured data.

\begin{figure}[ht!]
    \centering
    \includegraphics[width=0.6\textwidth]{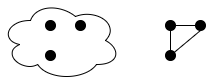}
    \caption{A hyperedge of size 3 (left) and a clique with 3 vertices (right).}
    \label{fig:clique_expansion}
\end{figure}

\paragraph{Clique expansion}
is an approach in which each hyperedge of size $k$ is substituted with $k*(k-1)$ edges amongst pairs of its members~\cite{sun2008hypergraph}. This approach cannot distinguish between pairwise relations amongst $k$ vertices and a hyperedge of size $k$, as shown in Figure~\ref{fig:clique_expansion}. In fact, they are both mapped to the same structure, and therefore lose structural information.

\begin{figure}[ht!]
    \centering
    \includegraphics[width=0.6\textwidth]{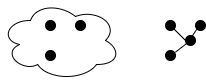}
    \caption{A hyperedge of size 3 (left) and its star expansion (right).}
    \label{fig:star_expansion}
\end{figure}

\paragraph{Star expansion}
is another approach of encoding a hypergraph in a simple graph, in which the hyperedge $h$ is replaced with a new vertex $v_h$, and edges between $h_v$ and original vertices of $h$ are added~\cite{zien1999multilevel}. An example could be seen in Figure~\ref{fig:star_expansion}. The resulting graph is bipartite, with one partition representing the original hyperedges and the other partition representing the original vertices. This conversion is not one to one because without further encoded information it is not possible to figure out which partition is representing the hyperedges. Another downside is that due to change in neighborhoods, in order to pass messages from previously neighbouring vertices, we now have to pass messages twice to reach the same destination.

\begin{figure}[ht!]
    \centering
    \includegraphics[width=0.6\textwidth]{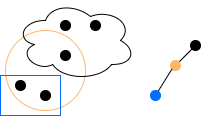}
    \caption{A hypergraph (left) and its line conversion (right).}
    \label{fig:line_conversion}
\end{figure}

\paragraph{Line conversions} (also known as edge conversion and edge expansion) creates a simple graph, and for any hyperedge $h$ creates a vertex $v_h$ in it. Two vertices $v_h$ and $v_h'$ are connected if and only if the intersection between $h$ and $h'$ is not empty~\cite{pu2012hypergraph}. 
An example is shown in Figure~\ref{fig:line_conversion}.
Plain line conversions are not a one to one mappings, and hence we loose various kind of information, from data stored on the vertices, to size of the higher order relations, neighbourhoods and many other types of structural information.

% \sajjad{the rest should be brief}
% To make up for it, there are two approaches that build on top of the line conversion, first one is the augmented edge, in which the edges in the output store what vertices were in the intersection of the two neighbouring hyperedges. This approach is not a one to one mapping either, and the loss of information on vertices still happens for any vertex that belongs to one or zero hyperedges.

% Finally, any hyperedge $h$ can store a concatenated list of the vertex features $X_u$, for all $u \in h$ on it's representative vertex $v_h$. To make things easier, the list could be zero padded to make them all the same size. The problem with this representation is that for disconnected hypergraphs, this approach is not one to one, and for connected ones, the processing model has to deal with an exceedingly large feature set on the graph vertices, increasing the number of learnable parameters.

\section{The Proposed Hypergraph Message Passing Neural Networks}
\label{sec:hmpnn}

The computation in the proposed Hypergraph Message Passing Neural Network (HMPNN) consists of two main phases: (1) sending messages from vertices to hyperedges and (2) sending messages from hyperedges to vertices. The operations performed by the proposed HMPNN model can be formalized as follows:
\begin{align}
    \label{hmpnn1}M_v &= f_v(X^{(t)}_v) \\
    \label{hmpnn2}W^{(t+1)}_e &= g_w(W^{(t)}_e, \square_{v \in e}M_v)\\
    \label{hmpnn3}M_e &= f_w(W^{(t)}_e, \square'_{v \in e}M_v)\\
    \label{hmpnn4}X^{(t+1)}_v &= g_v(X^{(t)}_v, \diamond_{e, v \in e}M_e)
\end{align}
$X^{(t)}_v$ is the representation of vertex $v$ at layer $t$, $W^{(t)}_e$ is the representation of hyperedge $e$ at layer $t$, $f_v$ and $f_w$ are vertex and hyperedge messaging functions, respectively, $g_v$ and $g_w$ are vertex and hyperedge updating functions, respectively, and finally $\square, \square'$, and $\diamond$ are aggregating functions.

The choice for messaging and updating functions includes identity function, linear and non-linear functions, or MLP. The choice for aggregation functions includes mean aggregation, sum aggregation and concatenation. 

We can therefore define an instance of hypergraph message passing operation by specifying what follows:
\begin{itemize}
    \item Choice of messaging functions over vertices $f_v$ and hyperedges $f_w$
    \item Choice of updating function over vertices $g_v$ and hyperedges $g_w$
    \item Choice of aggregation operation over incoming messages to hyperedges $\square$
    \item Choice of aggregation operation over outgoing messages to vertices $\square'$
    \item Choice of aggregation operation over incoming messages to vertices $\diamond$
\end{itemize}

These choices allow us to describe a wide range of hypergraph convolutions, which can be used as an intra-layer design space for hypergraphs, in a similar fashion to \citet{you2020design} for graphs.

It is possible to include various forms of regularization to prevent overfitting.
Dropout could be added to any of messaging or updating functions $f$ and $g$. It is also possible to introduce adjacency connection dropout to aggregation functions $\square \square'\diamond$, allowing for a reduction of bias during training.

\subsection{Batch Normalization and Dropout}

Batch normalization is a useful operation to speed-up training and reduce bias.
Similarly, dropout is a mechanism used to decrease the learned bias of the system. 
We introduce two types of dropout for HMPNN: regular dropout and adjacency dropout. During regular dropout, random cells in the feature vector of nodes or hyperedges are set to zero; the other cells are scaled by $\frac{n+k}{n}$, where $n$ is the total number of cells in that vector and $k$ is the number of cells that were zeroed. Adjacency dropout randomly removes hyperedges in a convolution step.

Adjacency dropout must be applied in neighborhood creation steps of Equations~\ref{hmpnn2} through \ref{hmpnn4}. Regular dropout can follow a batch normalization right before updating functions in Equations~\ref{hmpnn2} and \ref{hmpnn4}, as part of the corresponding $g$ functions.

\subsection{Equivalency between hypergraph convolutions and HMPNN}
\label{sec:}
Following the spirit of the graph design space mentioned in Section \ref{sec:graph_design_space}, here we show that, by including appropriate search parameters, HMPNN can mimic the behaviour of existing hypergraph neural network convolutions, which are thus special cases of HMPNN.

In order to compare different models based on hypergraph convolutions, we need to define a notion of equivalency among models. We define two models $m_1, m_2$ to be equivalent if and only if for any weights $w_1$ there exists $w_2$ such that for any input hypergraph $x$ the model outputs correspond, i.e. $m_1(w_1, x) = m_2(w_2, x)$; in other words we can translate the weights of one network to the other one.

\paragraph{Dynamic hypergraph neural network's convolution} (DHNN)~\cite{jiang2019dynamic} operates according to algorithm~\ref{alg:DHNN}.
In other words, for each hyperedge containing $u$, they sample all members of $u$ and capture their feature set (1st outgoing message), stacking them in $X_u$ (hyperedge aggregation), processing them via the vertex convolution(hyperedge updating), aggregating all such hyperedge messages via stacking (incoming messages to vertices), performing edge convolution on the stacked aggregation and applying the activation function on it (vertex update function).
\begin{algorithm}
\caption{Dynamic Hypergraph Neural Network Convolution}
\label{alg:DHNN}
\begin{algorithmic}
    \State \textbf{input}: Sample $x_u$, hypergraph structure $\mathcal{G}$
    \State \textbf{output}: Sample $y_u$
    \State xlist = $\Phi$
    \For e in Adj(u) do
        \State $X_u = \text{VertexSample}(X, \mathcal{G})$
        \State $X_e = \text{1-dconv}(MLP(X_v), MLP(X_v))$ or VertexConvolution($X_v$)
        \State xlist.insert($X_e$)
    \EndFor
    \State $X_e$ = stack(xlist)
    \State $X_u = \sum_{i=0}^{|Adj(u)|} \text{softmax}(x_eW+b)^i x^i_e$ or edgeConv($X_e$)
    \State $y_u = \sigma(x_uW+b)$
\end{algorithmic}
\end{algorithm}

These steps can be directly described in terms of message passing, with specific functions for each phase of message passing opearation. HMPNN can thus directly emulate this model.

\paragraph{Hypergraph neural networks} (HNN)~\cite{feng2019hypergraph}
convolution performs the following operations:
\begin{equation}
    X^{(l+1)} = \sigma(D_v^{-1/2}HWD_e^{-1}H^TD_v^{-1/2}X^{(l)}\Theta^{(l)})
\end{equation}
where $X^{(l)}$ describes the node feature on layer $l$, $D_v$ and $D_e$ describe the diagonal matrices of node and edge degrees, respectively, $W$ is the weights of each hyperedge, $H$ is the incidence matrix, $\Theta$ are the learnable parameters of the layer and finally $\sigma$ is the non-linear activation function. Note that this convolution assumes that hyperedge features are numbers (weights).

Given any $\Theta$, we construct the following equivalent HMPNN 
where the outgoing node message is node features multiplied by inverse square root of their degree, i.e. $D_v^{-1/2}X^{(l)}$; the hyperedge aggregation is the average, i.e. $D_e^{-1}H^T$; there is no hyperedge updating; the hyperedge outgoing message is multiplied by their weight (or feature); the node aggregation function is sum of input multiplied by the inverse square root of their degree, i.e. $D_v^{-1/2}H$; and finally the node updating function is $\sigma(X\Theta^{(l)})$.

\paragraph{Multiset Learning for hypergraphs} (AllSet)~\cite{chien2021you} construct their message passing according to the following equations:
\begin{align}
    &f_{\mathcal{V}\rightarrow\mathcal{E}}(S) = f_{\mathcal{E}\rightarrow\mathcal{V}}(S) = LN(Y+MLP(Y)),\\
    &Y= LN(\theta+MH_{h,\omega}(\theta,S,S)), MH_{h,\omega}(\theta,S,S)=||_i=1^{h} O^{(i)},\\
    &O^{(i)}= \omega(\theta^{(i)}(K^{(i)})^T)V^{(i)}, K^{(i)}=MLP^{(K,i)}(S), V^{(i)}=MLP^{(V,i)}(S).
\end{align}
$LN$ is the normalization layer, $||$ is concatenation and $\theta$ are the learnable weights, $MH_{h,w}$ denotes an h-multihead with activation function $\omega$. It is clear that the AllSet is a special case of the proposed HMPNN with the above mentioned functions used as the updating mechanism and the identity function used for messaging. Moreover, we note that there is a direct translation of weights from a trained AllSet model to the equivalent HMPNN.

\section{Experiments}
\label{sec:experiments}

In this section, we look at the semi-supervised task of node classification in citation networks, which offer a suitable playground to benchmark graph and hypergraph representations and related processing methods.

\subsection{Experiment Setup}

\subsubsection{Dataset}. To provide direct comparison with state-of-the-art methods \cite{kipf2016semi,bai2019hypergraph,velivckovic2017graph,feng2019hypergraph,jiang2019dynamic}, we use the Cora dataset\cite{yang2016revisiting} which includes 2708 academic papers and 5429 citation relations amongst them. Each publication contains a bag of word feature vector, where a binary value indicates presence or absence of that word in the publication. Finally, each publication belongs to one of the 7 categories.

For train-test split, we follow the protocol described in \cite{bai2019hypergraph,yang2016revisiting}. We randomly select: 20 items from each category, totaling 140 for training; 70 items from each category, totaling 490 for validation; the remaining items are used for testing.

\subsubsection{Hypergraph Construction}. Citation networks are often represented as simple graphs with publications that cite each other connected via simple edges. This representation fails to appreciate documents that are cited together, and only captures them in terms of a second neighborhood. \citet{bai2019hypergraph} provide an alternate representation in which documents are simultaneously treated as vertices and hyperedges at the same time, with hyperedges of a publication grouping all of its related work together. Each vertex is then characterized by a boolean matrix of their representative words acting as its feature set.

\subsubsection{Implementation Details}

Our model uses two layers of HMPNN with sigmoid activation and a hidden representation of size 2. We use sum as the message aggregation functions, with adjacency matrix dropout with rate 0.7, as well as dropout with rate 0.5 for vertex and hyperedge representation.

\subsection{Results and Analysis}

Table~\ref{tab:cora} shows a comparison with respect to different models taken from the state-of-the-art.
As shown in the table, the proposed method (HMPNN) significantly outperforms all other methods.

\begin{table}[htp!]
    \caption{Classification accuracy on Cora dataset.}
    \label{tab:cora}
    \centering
    \begin{tabular}{|c|c|}\hline
        \textbf{Method} & \textbf{Accuracy} \\\hline \hline
        GCN \cite{kipf2016semi} & 81.5\\\hline
        HGNN \cite{feng2019hypergraph} & 81.60\\\hline
        GAT \cite{velivckovic2017graph} & 82.43\\\hline
        DHGNN \cite{jiang2019dynamic} & 82.50\\\hline
        HGC+Atten \cite{bai2019hypergraph} & 82.61 \\\hline
        \textbf{HMPNN} & \textbf{92.16} \\\hline
    \end{tabular}
\end{table}

\subsubsection{Analysis of adjacency dropout}

In order to verify the benefits of adjacency dropout, we perform a test with various different values of adjacency dropout rate used during training and report the obtained accuracy on test set.
The results are shown in Table~\ref{tab:adj_dropout}.
Results show that, regardless of the use of activation dropout, the accuracy increases when we introduce adjacency dropout. We conclude that adjacency dropout is a useful mechanism to decrease the bias of the proposed neural network model.
\begin{table}[htp!]
\caption{Accuracy test for various values of adjacency dropout. Test 1 had 0.5 activation dropout whereas test 2 did not have any activation dropout.}
\label{tab:adj_dropout}
    \centering
    \begin{tabular}{|c|c|c|}\hline
        Adjacency Dropout Rate &  Accuracy Test 1 & Accuracy Test 2 \\\hline\hline
        90\% & 79.35\% & 65.51\%\\\hline
        70\% & 92.16\% & 89.90\%\\\hline
        50\% & 91.14\% & 83.26\%\\\hline
        30\% & 89.94\% & 76.18\%\\\hline
         0\% & 81.76\% & 54.11\%\\\hline
    \end{tabular}
\end{table}

\section{Conclusion and Future Work}
\label{sec:conclusions}

We investigate the use of hypergraph representations in machine learning and showed their superiority over graph representations with respect to capturing structural information for data characterized by $n$-ary relations (i.e. relations involving two or more vertices). Moreover, we introduced the Hypergraph Message Passing Neural Network as a novel neural network layer to process hypergraph-structured data. We showed that HMPNN can emulate existing convolution methods for hypergraphs, and therefore could be used to search over the space of architectures for hypergraph-structured data. 
Finally, we investigated using HMPNN on the task of node classification over the Cora citation network, in which we employed the adjacency dropout as well as activation dropout as mechanisms of controlling the bias. Our model outperformed various state-of-the-art methods for this task.

\bibliographystyle{splncs04nat}
\bibliography{bib}

\end{document}